\documentclass[preprint,12pt]{elsarticle}

\usepackage{amssymb}
\usepackage{amsmath}
\usepackage{hyperref}
\usepackage{color}

\usepackage{booktabs}
\usepackage{rotating}

\usepackage{array} 

\usepackage{graphicx}
\usepackage{pifont}
\newcommand{\cmark}{\ding{51}} 
\newcommand{\xmark}{\ding{55}} 

\begin{document}

\begin{frontmatter}



\title{A Multi-Stage Auto-Context Deep Learning Framework for Tissue and Nuclei Segmentation and Classification in H\&E-Stained Histological Images of Advanced Melanoma}

\author[label1]{Nima Torbati\fnref{equal}}
\author[label2,label3]{Anastasia Meshcheryakova\fnref{equal}}
\author[label1]{Ramona Woitek}
\author[label4,label5]{Sepideh Hatamikia}
\author[label2,label3]{Diana Mechtcheriakova\fnref{equal2}}
\author[label1]{Amirreza Mahbod\fnref{equal2}}

\fntext[equal]{Equal contributions}
\fntext[equal2]{Equal contributions}

\affiliation[label1]{organization={Research Center for Medical Image Analysis and Artificial Intelligence, Department of Medicine, Faculty of Medicine and Dentistry, Danube Private University},
            city={Krems an der Donau},
            postcode={3500},
            country={Austria}}

 \affiliation[label2]{organization={Department of Pathophysiology and Allergy Research, Center of Pathophysiology, Infectiology and Immunology, Medical University of Vienna},
             city={Vienna},
             postcode={1090},
             country={Austria}}

 \affiliation[label3]{organization={Comprehensive Center for AI in Medicine (CAIM), Medical University of Vienna},
             city={Vienna},
             postcode={1090},
             country={Austria}}
             
 \affiliation[label4]{organization={Research Center for Clinical AI-Research in Omics and Medical Data Science, Department of Medicine, Faculty of Medicine and Dentistry, Danube Private University},
	city={Krems an der Donau},
	postcode={3500},
	country={Austria}}
	
 \affiliation[label5]{organization={Austrian Center for Medical Innovation and Technology},
	city={Wiener Neustadt},
	postcode={2700},
	country={Austria}}
\date{}


\begin{abstract}
Melanoma is the most lethal form of skin cancer, with an increasing incidence rate worldwide. Analyzing histological images of melanoma by localizing and classifying tissues and cell nuclei is considered the gold standard method for diagnosis and treatment options for patients. While many computerized approaches have been proposed for automatic analysis, most perform tissue-based analysis and nuclei (cell)-based analysis as separate tasks, which might be suboptimal.

In this work, using the PUMA challenge dataset, we propose a novel multi-stage deep learning approach by combining tissue and nuclei information in a unified framework based on the auto-context concept to perform segmentation and classification in histological images of melanoma. Through pre-training and further post-processing, our approach achieved second and first place rankings in the PUMA challenge, with average micro Dice tissue score and summed nuclei F1-score of 73.40\% for Track 1 and 63.48\% for Track 2, respectively. Furthermore, through a comprehensive ablation study and additional evaluation on an external dataset, we demonstrated the effectiveness of the framework components as well as the generalization capabilities of the proposed approach. Our implementation for training and testing is available at:\url{https://github.com/NimaTorbati/PumaSubmit}
\end{abstract}




\end{frontmatter}

\section{Introduction}\label{sec1}
Skin cancer is one of the most common types of cancer, with melanoma being the most aggressive form~\cite{leiter2014epidemiology}. Although many non-invasive approaches such as dermatoscopic or clinical image analysis are widely used for early detection~\cite{diagnostics13071314,9858890,9412307}, examination through biopsy and subsequent analysis of histological images remains the gold standard for diagnosis~\cite{Elmorej2813}.

Various staining methods exist for histological specimens, but hematoxylin and eosin (H\&E) staining is the most commonly used by pathologists~\cite{deHaan2021}. To distinguish between primary melanomas (referred to as primary in the rest of the text) and distant metastases arising from melanomas (referred to as metastatic in the rest of the text), for example, medical experts and pathologists visually examine H\&E-stained slides and, based on their knowledge and experience, make diagnostic calls that guide treatment decisions. Several factors influence diagnostic accuracy, including the complexity of tissue anatomy and the cell type composition in the investigated specimens. Microscopic examination and assessment of histological tissue specimens by pathologist are labor-intensive and subject to intra- and inter-observer variability~\cite{10.1093/gigascience/giaf011}.

With the advancement of machine learning and deep learning (DL), automated computer-based methods have become powerful tools for performing tissue and nuclei segmentation and classification~\cite{Nasir2023}. These models can be trained using various approaches, including unsupervised, semi-supervised, self-supervised, and recently developed contrastive learning methods used in foundation model training~\cite{Qu_2022,bentham:/content/journals/cmir/10.2174/1573405617666210127154257,MAHBOD2025110571}. However, the most accurate models are typically trained in a supervised manner, which requires annotated data.

For supervised training of tissue and nuclei segmentation and classification models, several publicly available datasets exist. Datasets such as BACH~\cite{aresta2019bach}, GlaS~\cite{sirinukunwattana2017gland}, and CRAG~\cite{ALEMIKOOHBANANI2020101771} can be used for tissue segmentation. For nuclei segmentation and classification, datasets like PanNuke~\cite{10.1007/978-3-030-23937-4_2}, NuInsSeg~\cite{Mahbod2024}, CryoNuSeg~\cite{mahbod2021cryonuseg_org}, MoNuSAC~\cite{9446924}, and MoNuSeg~\cite{monuseg} can be utilized. 

In terms of model development, various approaches have been proposed for each task. Convolutional neural networks (CNNs)-based and vision transformer (ViT)-based models such as U-Net~\cite{Ronneberger2015}, U-Net++~\cite{10.1007/978-3-030-00889-5_1}, SegFormer~\cite{xie2021segformer,10750697}, and nnU-Net~\cite{isensee2021nnu,pmlr-v227-spronck24a} have been used for tissue segmentation in histological images. On the other hand, HoverNet~\cite{graham2019hover}, HoVer-NeXt~\cite{pmlr-v250-baumann24a}, CellViT~\cite{HORST2024103143}, CellViT++~\cite{horst2025cellvit++}, and DDU-Net~\cite{10.3389/fmed.2022.978146, mahbod2022deep} are examples of models developed specifically for nuclei instance segmentation and classification.

While these datasets and models have contributed significantly to the field of computational pathology, most of them treat tissue and nuclei segmentation as independent tasks, which may be suboptimal. Understanding the relationship between tissue context and nuclear characteristics has been shown to improve model performance for nuclei and tissue segmentation tasks~\cite{Ryu_2023_CVPR,Liu2024}. Furthermore, in specific clinical applications such as tumor-infiltrating lymphocyte (TIL) analysis, medical experts examine both tissue and nuclear features to support various diagnostic and prognostic assessments~\cite{van2021hooknet,Liu2024,schuiveling2025novel,van2025tumor,cancers12113117}.

The developed models for joint segmentation of tissue and nuclei typically consist of two components: one for tissue and one for nuclei. These components can be configured to exchange information either in parallel~\cite{van2021hooknet,Liu2024} or in series~\cite{millward2023dense,li2023enhancing,schoenpflug2023softctm}. In methods with a parallel structure, the tissue and nuclei networks are trained simultaneously, and information fusion between them can be achieved through parallel encoders with interconnections~\cite{shui2025towards}, parallel decoders~\cite{GRAHAM2023102685}, or parallel models utilizing specialized skip connections within the decoder~\cite{Liu2024}. In methods with a serial structure, a model such as SegFormer~\cite{xie2021segformer}, DeepLabv3+~\cite{chen2017rethinking}, or a ViT-based model~\cite{li2023enhancing} is first trained for the tissue segmentation task. The obtained results, together with the original input image, are then used to perform nuclei segmentation. To train such models, a few publicly available datasets for various tissue types, such as TIGER, PanopTILs, OCELOT, and the recently published PUMA dataset, can be used. While these models have shown improved performance for tissue and nuclei segmentation tasks, further improvements are still needed. Moreover, their application to histological image analysis of advanced melanoma has not yet been investigated.

For DL-based melanoma histological image analysis, most developed methods focus on distinct tasks of nuclei and tissue segmentation~\cite{mosquera2022deep}, and only a few approaches have attempted to integrate both nuclei- and tissue-level information. Liu et al.~\cite{liu2021learning} investigated the performance of the Mask R-CNN network~\cite{he2017mask,pmlr-v156-bancher21a} for detecting potential melanoma regions using a sparse and noisy dataset annotated with both tissue and nuclei labels. However, training a model on such data without a systematic strategy for integrating tissue and nuclei information may be suboptimal. In another study~\cite{alheejawi2021detection}, a method was proposed for melanoma region detection based solely on nuclei segmentation. This approach first employs a CNN-based model to perform nuclei segmentation, followed by binary morphological operations to group nuclei into melanoma regions. However, relying solely on nuclei-level features limits the model’s ability to capture global contextual dependencies. Moreover, the performance of such binary approaches is often sensitive to hyperparameters, such as the size of the dilation kernel. In~\cite{parajuli2023automated}, a weakly supervised framework was introduced to differentiate skin melanocytes from keratinocytes. This approach initially trains a DeepLabv3+ network to segment epidermal tissues. Patches extracted from these regions are then processed using a pre-trained VGG16-based classifier~\cite{Simonyan2014}. To differentiate melanocytes from keratinocytes, the method applies Otsu thresholding~\cite{4310076} to the gradient maps of the final classifier layer (similar to Grad-CAM~\cite{selvaraju2017grad}). While this method avoids the need for nuclei annotations, it is not readily applicable to multi-class or instance segmentation tasks, limiting its scalability and generalizability.

Considering the current limitations of existing models for joint nuclei and tissue segmentation in histological images of advanced melanoma, there is a clear need for a robust and generalizable framework to address these tasks. Another major limitation is that most of these models have been developed using private datasets, which restricts their reproducibility and broader applicability. To address the latter issue, the Panoptic segmentation of nUclei and tissue in advanced MelanomA (PUMA) challenge dataset~\cite{10.1093/gigascience/giaf011} was recently introduced. This dataset contains expertly annotated images of both tissue and nuclei, and it may serve as a valuable benchmark for research specifically focused on melanoma histological image analysis. The PUMA challenge introduces a unique dataset and encourages the research community to develop robust models for two main tasks defined in its two tracks. Track 1 involves segmentation and classification of five tissue types and three nuclei instance types.
Track 2 includes segmentation and classification of the same five tissue types and ten nuclei instance classes. 

In this study, using the PUMA challenge dataset, we propose a novel multi-stage approach based on the auto-context concept~\cite{tu2009auto}. The proposed method consists of four sequential stages: classification, initial tissue segmentation, nuclei segmentation, and final tissue segmentation. In summary, in stage 1, a classifier is designed to enhance model performance by leveraging the biological characteristics of tissue structures. In stage 2, global contextual information is extracted through the segmentation of tissues. In stage 3, the tissue segmentation output from stage 2 is fused with the input image to guide nuclei segmentation. Finally, in stage 4, the nuclei segmentation result from stage 3 is fused with the input image to perform the final tissue segmentation. The results on the final test set of both tracks of the challenge confirmed the excellent performance of our method, achieving a mean micro Dice tissue score and summed nuclei F1-score of 73.40\% for Track 1 and 63.48\% for Track 2, respectively. These results ranked our approach (submitted under the name "LSM team") second in Track 1 and first in Track 2 of the PUMA competition, respectively~\footnote{\url{https://puma.grand-challenge.org/}}. Through a comprehensive ablation study, we demonstrated the effectiveness of each stage of our proposed model on overall performance. Additionally, we evaluated the performance of our approach for nuclei instance segmentation and classification on an external dataset (the MoNuSAC challenge data), and showed improved results compared to the top performers of the challenge.

\section{Method}\label{sec1}

We propose a multi-stage auto-context pipeline, where the outputs of earlier stages are used as contextual input to refine tissue segmentation and nuclei segmentation and classification in later stages. Our method consists of four main stages, as illustrated in Figure~\ref{fig:stage1}, Figure~\ref{fig:stage2}, Figure~\ref{fig:stage3}, and Figure~\ref{fig:stage4}.

In the first stage (Figure~\ref{fig:stage1}), we designed a classifier to determine the input image type, distinguishing between primary and metastatic frames. In the second stage (Figure~\ref{fig:stage2}), initial tissue segmentation is performed. The third stage (Figure~\ref{fig:stage3}) involves nuclei instance segmentation using the baseline HoVer-NeXt model, along with nuclei classification, which utilizes the tissue information from the previous step. Finally, in the fourth stage (Figure~\ref{fig:stage4}), tissue segmentation is refined using the nuclei masks produced in the third stage. Further description of our pipeline is given below.

\subsection{Datasets}
\label{sec:dataset}

We used three datasets in this work: the PanopTILs dataset~\cite{Liu2024}, the PUMA dataset~\cite{10.1093/gigascience/giaf011}, and the MoNuSAC dataset~\cite{9745890}, each for different purposes.

The PanopTILs dataset was used exclusively for pre-training. The PUMA dataset served as the main dataset for training and evaluation, while the MoNuSAC dataset was used for external validation of our proposed approach.

The PanopTILs dataset focusing on breast tumor microenvironment contains a total of 1,709 image patches extracted from whole-slide images. Each patch has a size of 1024$\times$1024 pixels with an image resolution of 0.25 $\mu$m per pixel, which is approximately equivalent to 40$\times$ magnification. Since the PUMA dataset images have a slightly different resolution (0.23 $\mu$m per pixel), we first cropped the PanopTILs images to 880$\times$880 pixels (top-left cropping) and then resized them to 1024$\times$1024 pixels to ensure a similar resolution comparable to that of the PUMA images. The PanopTILs dataset includes annotations for nine tissue types (exclude, cancerous epithelium, stroma, TILs, normal epithelium, junk/debris (necrosis), blood vessel, other, and white space/empty) and 10 nuclei types (exclude, cancer nucleus, stromal nucleus, large stromal nucleus, lymphocyte nucleus, plasma cell / large TIL nucleus, normal epithelial nucleus, other nucleus, unknown/ambiguous nucleus, and background). From the original 1,709 image regions, we used only 300 images for pre-training and merged some classes to align the labels more closely with those used in the PUMA dataset. For tissue classes, we grouped “exclude,” “other,” and “white space/empty” as background. Additionally, “cancerous epithelium” was categorized as tumor, and “TILs” class was merged with stroma. Only images containing necrosis or tumor tissue were used in our analysis, while those containing the normal epithelium class were excluded.

The second dataset used in this study is the PUMA dataset, which served as the main dataset for our experiments. It contains a total of 310 melanoma image patches: 155 from primary tumors and 155 from metastatic tumors, all with a size of 1024$\times$1024 pixels. These were divided into training (206 images), validation (10 images), and test (94 images) sets. While both tissue and nuclei annotations were provided for the training images, the corresponding labels for the validation and test sets were not disclosed by the challenge organizers. Participants in the challenge could evaluate their models on the 10 validation images (also referred as preliminary test set in the challenge) through the PUMA Grand Challenge platform~\footnote{\url{https://puma.grand-challenge.org/}} and submit their final results for the 94 test images. The PUMA dataset includes annotations for six tissue types: tumor, stroma, epidermis, necrosis, blood vessel, and background. For nuclei, Track 1 of the challenge includes three types: tumor cells, lymphocyte, and other. Track 2 includes ten types: tumor cell, lymphocyte, plasma cell, histocyte, melanophage, neutrophil, stroma cell, endothelium, epithelium, and apoptotic cell, as well as the background class. 
In our experiments, we used all training images except one due to a mismatch between the image and the provided segmentation mask. Further details about the PUMA dataset can be found in~\cite{10.1093/gigascience/giaf011}.

As it is essential to demonstrate the generalization capability of the proposed pipeline~\cite{MAHBOD2024669}, we used the MoNuSAC~\footnote{\url{https://monusac-2020.grand-challenge.org/}} dataset as an external validation set. The MoNuSAC dataset comprises 310 image patches of varying sizes from H\&E-stained tissue sections. Specimens were derived from patients with lung, prostate, kidney, and breast cancer. The dataset consists of 209 training images and 101 test images. It includes annotations for four types of nuclei: epithelial cells, lymphocytes, macrophages, and neutrophils. However, tissue-level annotations are not provided.

It is worth noting that the flowcharts presented in the remainder of the manuscript illustrate only the fine-tuning steps using the PUMA dataset and do not depict the pre-training conducted with the PanopTILs dataset. The training and evaluation procedure for the MoNuSAC dataset is described separately in Section~\ref{sec:Training Procedure for MoNuSAC}.

\subsection{Utilized Backbone Models}
In this study, we used four state-of-the-art segmentation models at different stages of our proposed method. The first model is SegFormer~\cite{xie2021segformer} (B2 variant) with ImageNet pre-trained weights~\cite{Russakovsky2015}. The second model is U-Net~\cite{Ronneberger2015} with a pre-trained ResNet34 encoder~\cite{Yakubovskiy:2019}. The third model is UNet++~\cite{10.1007/978-3-030-00889-5_1} with a pre-trained ResNet50 encoder~\cite{Yakubovskiy:2019}. Finally, the fourth model is HoVer-NeXt, which was used exclusively for nuclei instance segmentation. The pre-trained HoVer-NeXt model, based on the PUMA training dataset, was provided by the challenge organizers. 

\subsection{Stage 1: Classifier}
To address the challenges posed by the limited dataset size and variations in tissue types, we differentiated tissue types between the primary tumors and metastatic sites. Our approach involves training of two separate models: one for primary images and another for metastatic images. Therefore, an initial classification step is required to determine the image type. 
To build the classifier, we first increased the number of tissue classes from five to ten, with the first five representing primary tissue types and the remaining five corresponding to metastatic tissue types. We then trained a SegFormer-B2 segmentation model with 11 classes (including the background) as the main classifier, followed by a set of handcrafted classification rules to categorize images as either primary or metastatic, as stated below.

Since epidermis is a skin-specific tissue, an image is classified as primary if the segmentation result includes "epidermis" class. If the epidermis is not present, the image is still classified as primary if the total number of primary pixels exceeds the total number of metastatic pixels. Otherwise, the image is classified as metastatic.

\begin{figure}[h]
	\centering
	\includegraphics[width=0.9\textwidth]{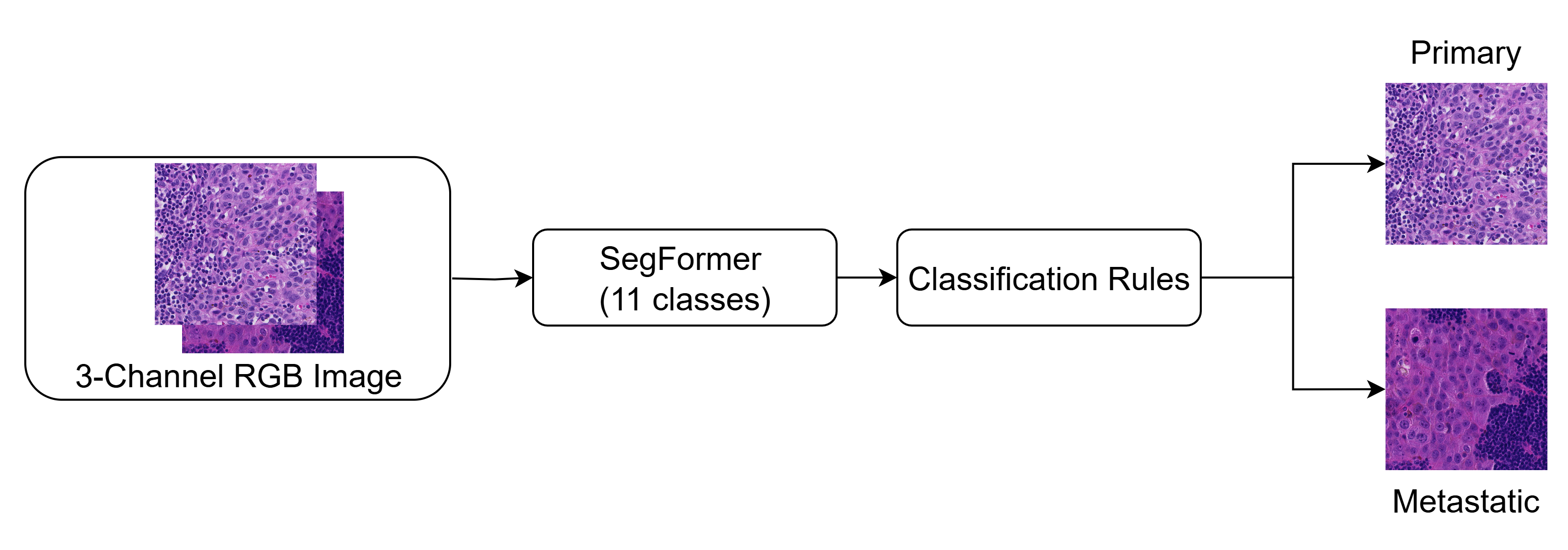}
	\caption{The first stage of the proposed method. In this stage, a SegFormer model is trained to classify the input image type (primary or metastatic) based on segmentation result  and the classification rules. 
	}\label{fig:stage1}
\end{figure}

The flowchart of this stage  is shown in Figure~\ref{fig:stage1}.

\subsection{Stage 2: Initial Tissue Segmentation}

In this stage, two SegFormer models are trained for initial tissue segmentation. One model is trained on the primary subset of the PUMA dataset, and the other is trained on the metastatic subset. Based on the classification results from stage 1, each input frame is directed to the appropriate model.

During our experiments, we observed that the performance of the SegFormer model in detecting blood vessels was suboptimal. To address this, we trained a U-Net model using all available training images in the PUMA dataset specifically for blood vessel detection. As a result, in our approach, we replaced the SegFormer predictions for blood vessels with the corresponding U-Net predictions.

To further enhance the segmentation performance, we defined a set of tissue ensemble rules. The final prediction for epidermis and necrosis is computed as the average of the predictions from both SegFormer and U-Net models:

\begin{equation}
	P_{\text{Epidermis}} = 0.5 \left( P_{\text{SegFormer\_Epidermis}} + P_{\text{UNet\_Epidermis}} \right)
\end{equation}

\begin{equation}
	P_{\text{Necrosis}} = 0.5 \left( P_{\text{SegFormer\_Necrosis}} + P_{\text{UNet\_Necrosis}} \right)
\end{equation}

\noindent For the blood vessel class, as described, we directly use the U-Net output.

\begin{equation}
	P_{\text{Blood vessel}} =  P_{\text{UNet\_Blood vessel}} 
\end{equation}

\noindent The predictions for tumor and stroma tissues are solely based on the SegFormer output:

\begin{equation}
	P_{\text{Tumor}} =  P_{\text{SegFormer\_Tumor}} 
\end{equation}

\begin{equation}
	P_{\text{Stroma}} =  P_{\text{SegFormer\_Stroma}} 
\end{equation}

\begin{figure}[h]
	\centering
	\includegraphics[width=0.9\textwidth]{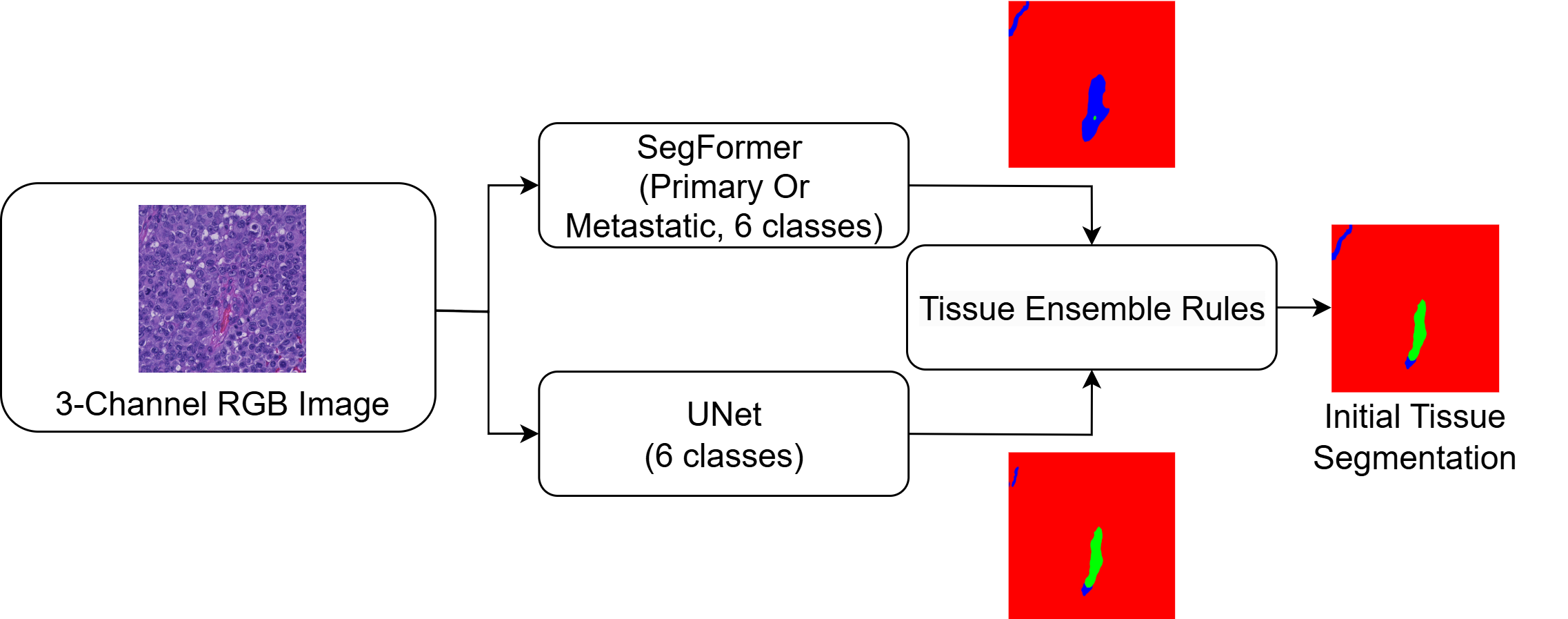}
	\caption{The second stage of the proposed method. Initial tissue segmentation is performed using two SegFormer models, one for primary tissues and the other for metastatic tissues. Due to the Segformer model's poor performance in segmenting blood vessels, a U-Net model is trained separately for blood vessel detection and the results are refined using the tissue ensemble rules. Color code for tissue segmentation: red for tumor, blue for stroma, and green for blood vessel.
	}\label{fig:stage2}
\end{figure}

The flowchart of this stage  is shown in Figure~\ref{fig:stage2}.


\subsection{Stage 3: Nuclei Segmentation}
\begin{figure}[h]
	\centering
	\includegraphics[width=1.0\textwidth]{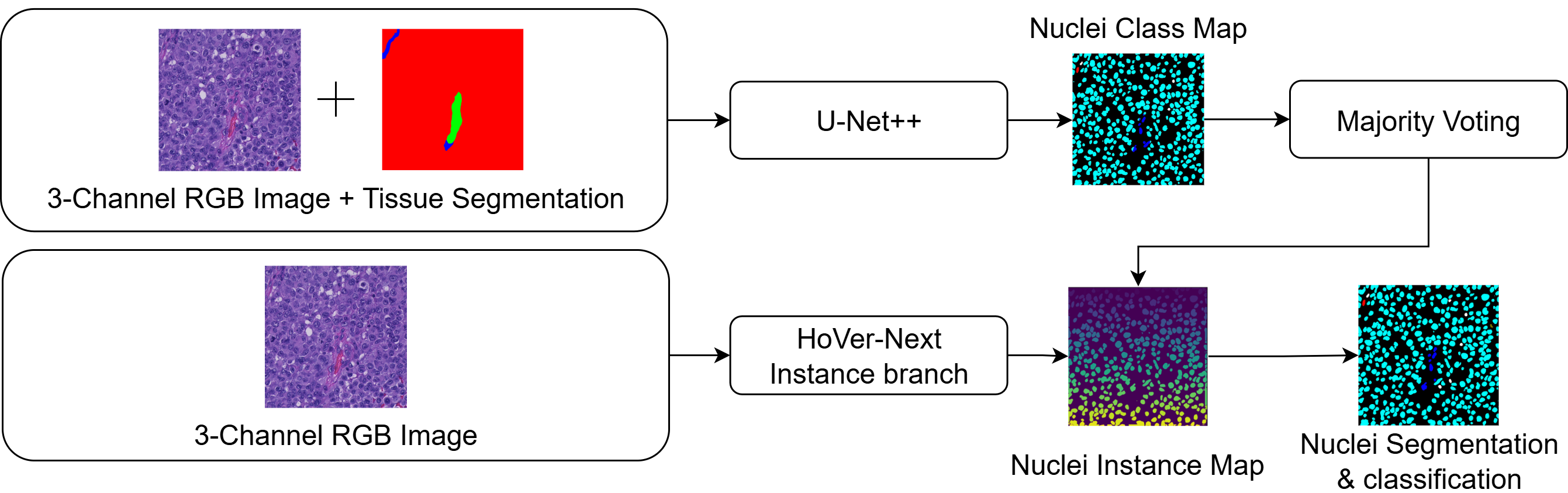}
	\caption{The third stage of the proposed method. This stage consists of two branches, with the upper part dedicated to nuclei classification and the lower part to nuclei instance segmentation. In the first branch, a U-Net++ model is trained for nuclei class map detection, incorporating tissue segmentation results as an additional input channel. In the second branch, nuclei instance segmentation is performed using the HoVer-NeXt model. The final nuclei segmentation and classification is obtained by combining the class maps and instance masks through a majority voting approach. Color code for tissue segmentation: red for tumor, blue for stroma, and green for blood vessel; color code for nuclei segmentation: various cell types are indicated by different colors.}\label{fig:stage3}
\end{figure}

As shown in Figure~\ref{fig:stage3}, this stage consists of two branches: one for nuclei class map prediction and the other for nuclei instance segmentation. In the first branch, we incorporate the tissue segmentation result from stage 2 as a fourth input channel and train a U-Net++ model for nuclei class map prediction. In the second branch, nuclei instance masks are generated using the pre-trained HoVer-NeXt model~\cite{pmlr-v250-baumann24a}, provided by the challenge organizers.

As observed in our experiments and inferred from the results in~\cite{10.1093/gigascience/giaf011}, rather than improving the instance segmentation outputs, we chose to use the baseline instance segmentation results from the HoVer-NeXt model and focused on enhancing the classification results using an external model. Specifically, we employed U-Net++ with four input channels for this task. We previously tested this approach (that is, using an external model solely for nuclei classification) in the MoNuSAC challenge, where we achieved first place on the post-challenge leaderboard~\cite{10.3389/fmed.2022.978146,9446924}. In this study, for each detected instance from the HoVer-NeXt model, we determined the instance class from the U-Net++ predictions using a simple majority voting approach. The number of U-Net++ output classes depends on the challenge track: four classes for Track 1 and eleven classes for Track 2, including the background in both cases.


\subsection{Stage 4: Final Tissue Segmentation}
\begin{figure}[h]
	\centering
	\includegraphics[width=1.0\textwidth]{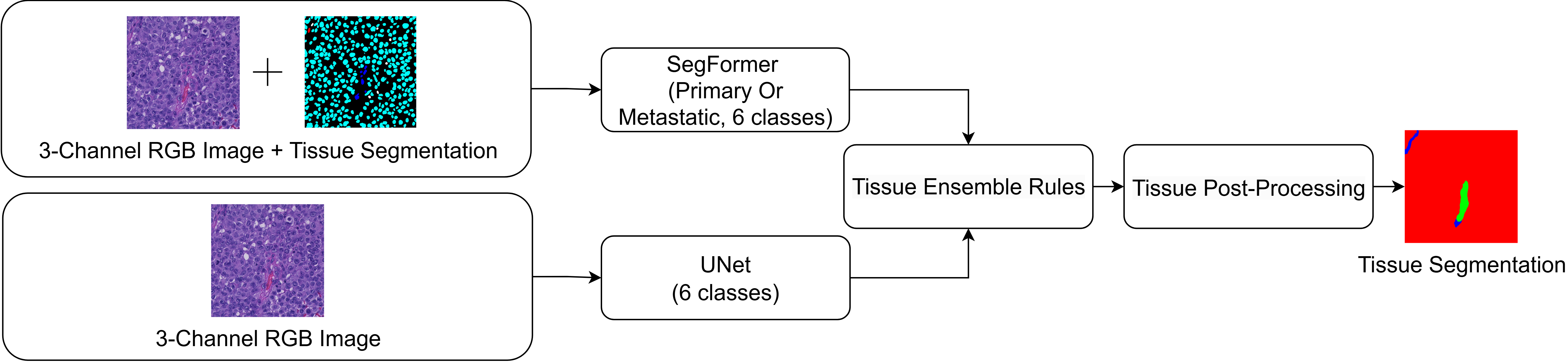}
	\caption{The fourth and final stage of the proposed method. Tissue segmentation is refined using nuclei segmentation results. The nuclei segmentation output from the previous stage is added as an additional input channel, and two SegFormer models similar to those in stage 2 are trained. For the U-Net model, we used the same trained model in stage 2 without incorporating the fourth nuclear channel. Color code for tissue segmentation: red for tumor, blue for stroma, and green for blood vessel; color code for nuclei segmentation: various cell types are indicated by different colors.}\label{fig:stage4}
\end{figure}

In this stage, we incorporate the nuclei segmentation result from stage 3 as a fourth input channel to train two SegFormer models similar to those in stage 2 as shown in Figure~\ref{fig:stage4}. We used the same U-Net trained model in stage 2 without incorporating the fourth nuclear channel, in order to simplify the framework while not degrading the overall performance, particularly for blood vessel detection. Tissue ensemble rules used in stage 4 are the same as in stage 2.


\subsection{Post-Processing}
\label{sec:postprocess}

To enhance the performance, we applied two post-processing steps to the tissue and nuclei segmentation results.

\subsubsection{Tissue Post-Processing}
\begin{figure}[t!]
	\centering
	\begin{tabular}{cccc}
		\includegraphics[width=4cm]{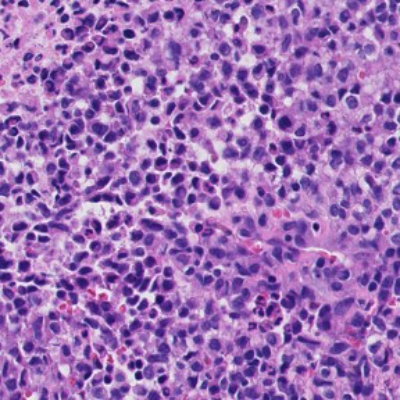} &
		\includegraphics[width=4cm]{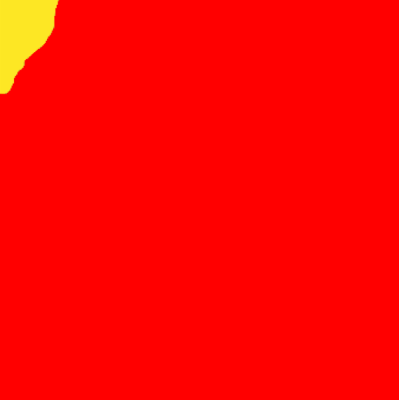} &
		\includegraphics[width=4cm]{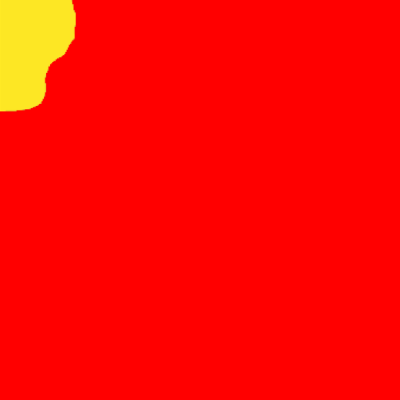}   \\
		
		(a)	   &  	(b) &	 (c)  \\
		
	\end{tabular}
	\caption{Post-processing of necrotic tissue. Due to the lower density of nuclei in necrotic part of the tissue (a), the final segmentation result in stage 4 may suffer from under-segmentation (b). To address this, a post-processing step integrates predictions from stage 1 when they overlap with stage 4 predictions for necrotic tissue (c). Color code in (b) and (c): yellow for necrotic part of the tissue, red for tumor.}
	\label{fig:post_necrotic}
\end{figure}

In stage 4, we observed under-segmentation of necrotic tissue. This issue arises due to the low density of nuclei in necrotic regions. An example of this under-segmentation is shown in Figure~\ref{fig:post_necrotic}. To mitigate this, we incorporate predictions from stage 1 when there is an overlap between the stage 1 and stage 4 predictions for necrotic tissue. We applied this post-processing step only in Track 2 of the challenge and not in Track 1. As the results confirm, this post-processing significantly improved the tissue segmentation performance, especially for necrosis (refer to Table~\ref{tab:track1}, Table~\ref{tab:track2}, and Table~\ref{tab:tissue_deatils} for further details).

\subsubsection{Nuclei Post-Processing}

\begin{figure}[t!]
	\centering
	\begin{tabular}{cccc}
		
		\includegraphics[width=4cm]{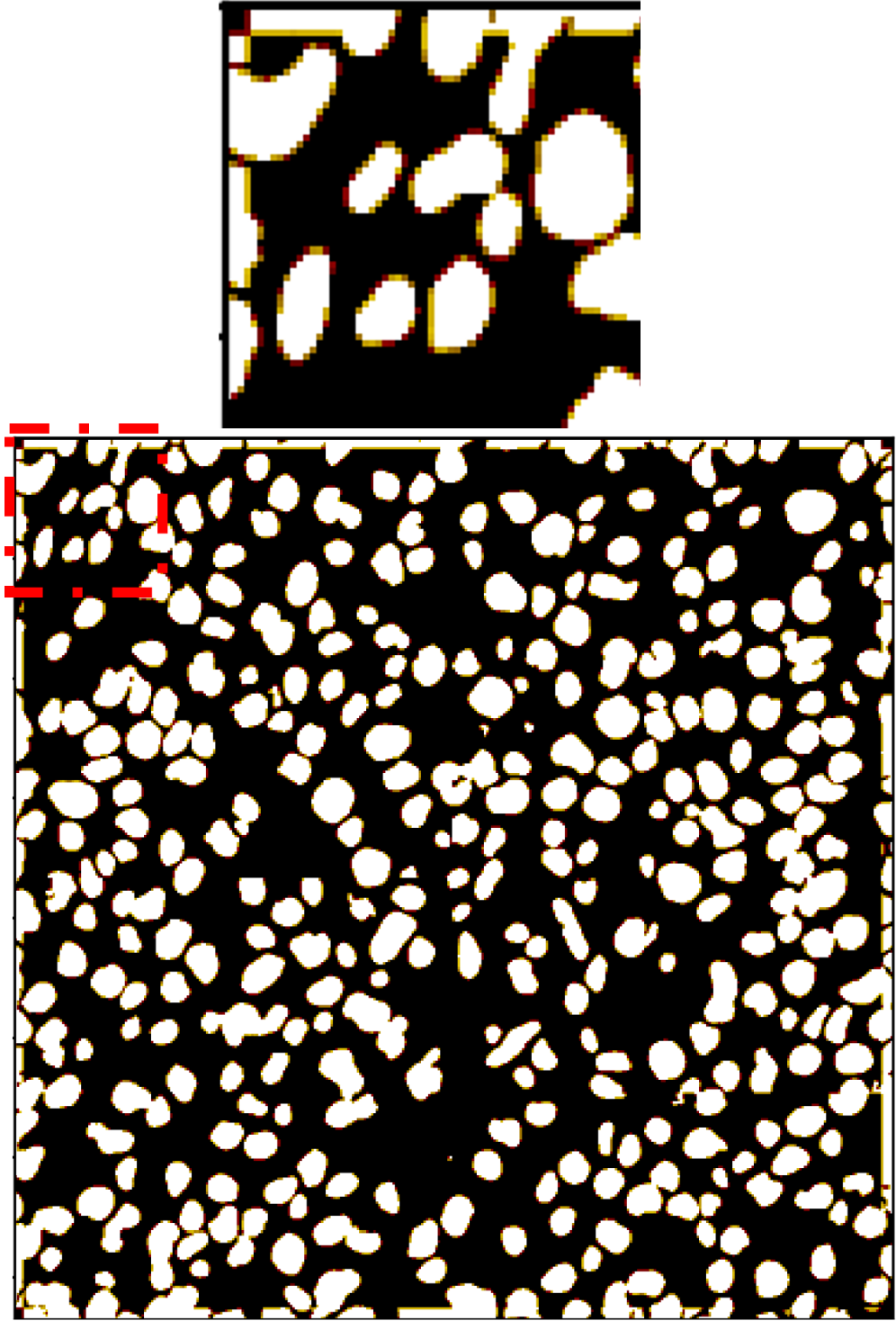} &
		\includegraphics[width=4cm]{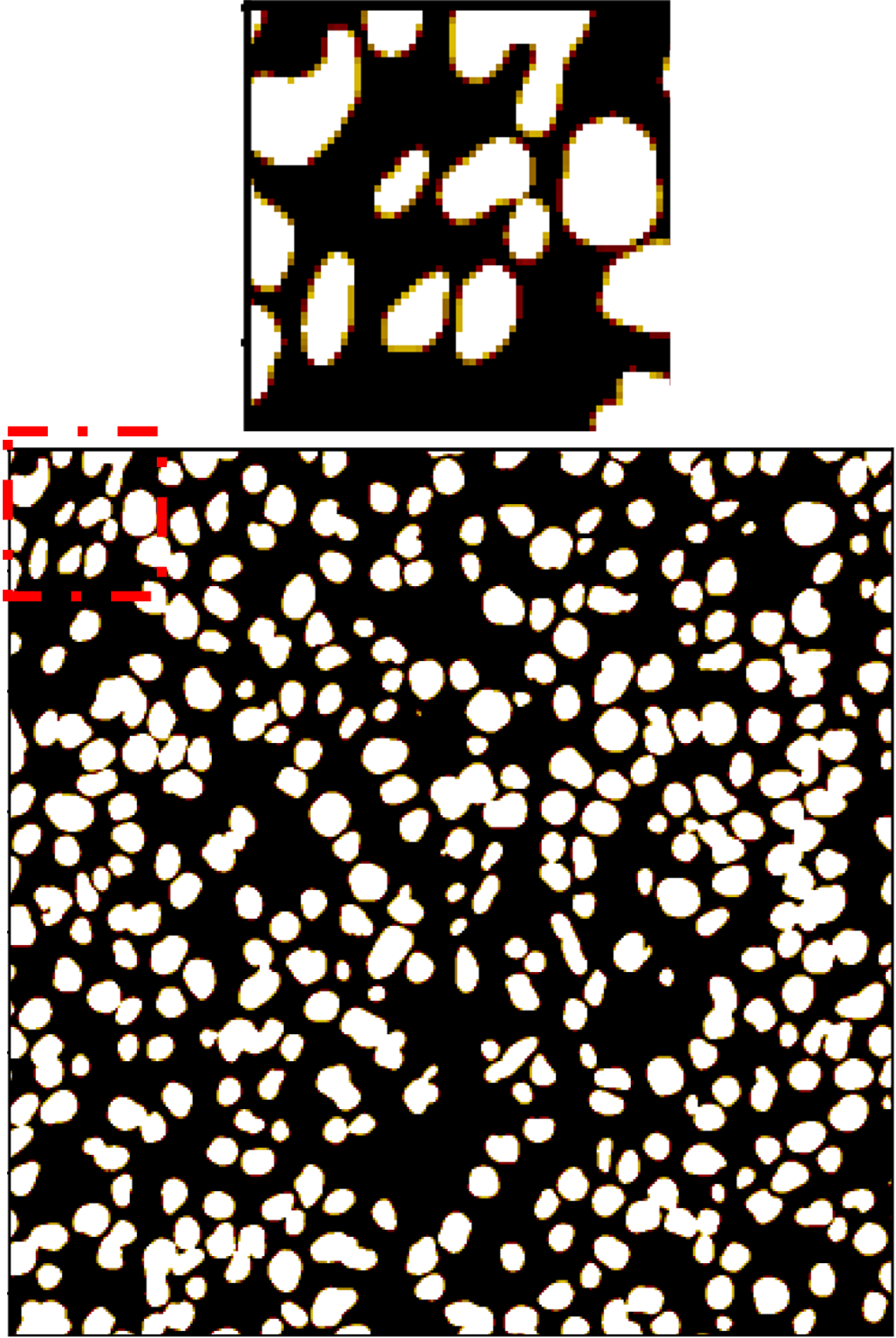} &
		\includegraphics[width=4cm]{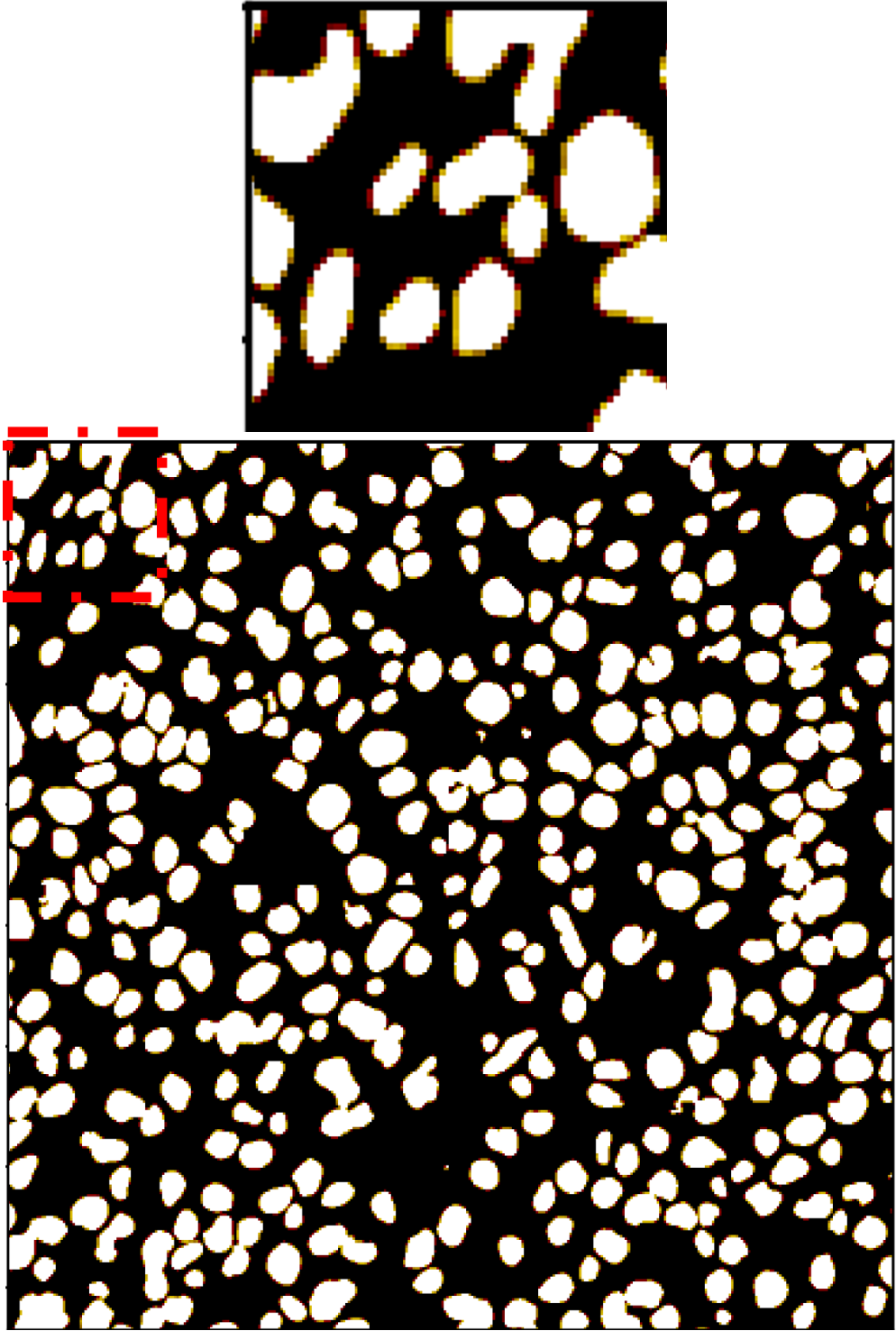}   \\
		
		(a)	   &  	(b) &	 (c)  \\
		
	\end{tabular}
	\caption{Post-processing at image borders. At image borders, nuclei instance segmentation results are replaced with class map results. (a) Binary segmentation map from HoVer-Next. (b) Binary class map from U-Net ++. (c) Binary map after border correction. Color code: white for nuclei, black as background. Exemplified differences are shown in red boxes.}
	\label{fig:post_border}
\end{figure}

For nuclei segmentation, corrections are required for the instance segmentation map at image borders. Since HoVer-NeXt divides the input image into small patches during the inference phase, its performance at image borders is suboptimal. An example of this issue is shown in Figure~\ref{fig:post_border}. To address this, we replace the instance segmentation results at image borders with the corresponding class map results.


\subsection{External Validation on the MoNuSAC Dataset}
\label{sec:Training Procedure for MoNuSAC}
The MoNuSAC dataset contains only nuclei annotations and does not include tissue-level annotations. Therefore, we adapted the proposed framework for this dataset as follows. Since skin tissue images are not present in the MoNuSAC data, Stage 1 was omitted, and all images were treated as one class and models trained with metastatic images were used. In Stage 2, we applied the pre-trained tissue segmentation model from the PUMA dataset to generate tissue predictions. These predictions were then used in Stage 3, where the nuclei segmentation model was trained using the MoNuSAC training images. For the instance segmentation branch in Stage 3, we separately used nuclei instance masks provided by the MoNuSAC challenge winner team~\cite{graham2019hover} and the post-challenge winner team~\cite{10.3389/fmed.2022.978146}, both of which are publicly available on the MoNuSAC challenge website. The performance of the proposed method was evaluated on the test images of the MoNuSAC dataset.

\subsection{Evaluation}
To evaluate the performance of the algorithms, the micro Dice score for multi-class tissue segmentation and the summed macro F1-score for nuclei instance detection and classification were used for the PUMA dataset as the official evaluation metrics of the challenge. The micro Dice score is calculated by treating all segmentation results from all images as a single large image. The summed macro F1-score is determined using a 15-pixel centroid threshold between the predicted segmentation of the nuclei results and the ground truth. For the final ranking of the PUMA challenge, the average of the micro Dice score and the summed macro F1-score was used for both Track 1 and Track 2. Further details and implementation of the utilized metrics are available in the PUMA challenge description\footnote{\url{https://puma.grand-challenge.org/submission/}}.

For the MoNuSAC dataset, we adopted the class-specific Panoptic Quality (PQ) score, consistent with the official MoNuSAC challenge evaluation protocol. Additionally, inspired by the micro Dice score, we calculated the micro PQ by treating all segmentation results from all images as a single large image. Further details and implementation of the PQ are available in the MoNuSAC challenge description\footnote{\url{https://monusac-2020.grand-challenge.org/Evaluation_Metric/}}.

\section{Experiments}

To conduct experiments, we utilized both pre-training with the PanopTILs dataset (as described in Section~\ref{sec:dataset}) and data augmentation techniques to improve the performance of the models. For data augmentation, we applied both shape-based transformations (flip, rotation, scaling) and intensity-based adjustments (hue, saturation, brightness, contrast). All experiments were conducted using a single workstation with AMD Ryzen 9 7950X CPU, 96 GB of RAM, and two NVIDIA RTX 4090 GPUs. Our implementation for training and testing is available on our GitHub repository: \url{https://github.com/NimaTorbati/PumaSubmit}.

\section{Results \& Discussion}\label{sec2}

This section presents the evaluation results of the proposed method on the PUMA and MoNuSAC datasets. Additionally, it includes an ablation study analyzing the contribution of each stage within the proposed framework.

\subsection{PUMA Dataset Results}
The final test results and rankings of Track 1 and Track 2 for the top three performers in the PUMA challenge, as well as the baseline results provided by the challenge organizers (nnU-Net~\cite{isensee2021nnu} for tissue segmentation and HoVer-NeXt~\cite{pmlr-v250-baumann24a} for nuclei segmentation and classification), are shown in Table~\ref{tab:track1} and Table~\ref{tab:track2}, respectively.

As the results show, all three top-ranked teams outperformed the provided baseline method by a large margin. Details of the methods used by the other top-performing teams among those TIAKong, rictoo, and agaldran are not available at the time of writing this manuscript but are expected to be released by the organizers on the challenge website. With our proposed approach (LSM), we achieved second place in Track 2 and first place in Track 1 of the competition.

When considering only the performance on nuclei instance segmentation and classification, we achieved the second-best result in Track 1 (behind the rictoo team with 75.77\%) and the best performance in Track 2. However, for multi-class tissue segmentation, our method ranked third in Track 1 (following TIAKong with 78.23\% and Biototem with 72.68\%) and second in Track 2 (after TIAKong with 78.23\%).

\begin{table*}
	\centering
	\caption{The final test results and ranking of Track 1 of the PUMA challenge (\%).}
	\label{tab:track1}
	\begin{tabular}{l@{\hskip 20pt}c@{\hskip 20pt}c@{\hskip 20pt}c@{\hskip 20pt}c}
		\hline
		Rank & Team & \begin{tabular}[c]{@{}c@{}}Micro Dice  \\  (Tissue)\end{tabular} & \begin{tabular}[c]{@{}c@{}}Summed macro F1  \\  (Nuclei) \end{tabular} & \begin{tabular}[c]{@{}c@{}}Mean\\ \end{tabular} \\
		\hline
		1    &  TIAKong & \textbf{78.23}   & 74.39           & \textbf{76.31} \\
		2    &  LSM     & 72.37            & 74.43           & 73.40          \\
		3    &  rictoo  & 63.26            & \textbf{75.78}  & 69.52          \\
		8    &  Basline & 55.48            & 69.40           & 62.44          \\
		\hline
	\end{tabular}
\end{table*}

\begin{table*}
	\centering
	\caption{The final test results and ranking of Track 2 of the PUMA challenge (\%).}
	\label{tab:track2}
	\begin{tabular}{l@{\hskip 20pt}c@{\hskip 20pt}c@{\hskip 20pt}c@{\hskip 20pt}c}
		\hline
		Rank & Team & \begin{tabular}[c]{@{}c@{}}Micro Dice  \\  (Tissue)\end{tabular} & \begin{tabular}[c]{@{}c@{}}Summed macro F1  \\  (Nuclei) \end{tabular} & \begin{tabular}[c]{@{}c@{}}Mean\\ \end{tabular} \\
		\hline
		1    &  LSM       & 77.98            & \textbf{48.97}  & \textbf{63.48} \\
		2    &  TIAKong   & \textbf{78.23}   & 46.69           & 62.46          \\
		3    &  agaldran  & 62.04            & 47.78           & 54.91          \\
		11   &  Baseline  & 55.48            & 29.77           & 42.63          \\
		\hline
	\end{tabular}
\end{table*}

As pointed out in Section~\ref{sec:postprocess}, we only applied the proposed tissue segmentation post processing in Track 2 of the challenge. As the average results indicate, this step significantly improved overall performance. 

Another observation from the results in Table 1 and Table 2 is the significant performance drop in nuclei classification and segmentation as the number of nuclei classes increases. For example, the summed macro F1 score for Track 1 (with three nuclei classes) using our proposed approach is 74.43\%, whereas for Track 2 (with ten nuclei classes), it drops to 48.97\%. This performance decrease could be attributed to the increased complexity of Track 2 compared to Track 1. Additionally, it may be related to the challenges of accurately annotating H\&E-stained images when a larger number of nuclei classes are involved. This limitation also has biological basis and is biologically relevant, as other staining techniques—such as immunohistochemistry (IHC)—are commonly applied to visualize individual cell types using specific cellular markers in histological images. Consequently, detection and classification of certain nuclei types based solely on H\&E staining may not be feasible. To address this challenge and achieve accurate nuclei segmentation and classification for specific cell types, using specially designed and annotated datasets based on marker-specific staining techniques could be a promising solution.

We also report the tissue-specific scores on the final test sets of Track 1 and Track 2 in Table~\ref{tab:tissue_deatils}. According to these results, the post processing step notably enhanced the segmentation performance for necrosis tissue (from 46.69\% to 74.49\%). However, it should be noted that necrotic part of tissue is the least represented tissue class across the training, validation, and test sets. Therefore, even small over- or under-segmentation can significantly affect the micro Dice score for necrosis, and thus influence the overall average tissue segmentation performance. In Table~\ref{tab:tissue_deatils}, we also compare our tissue-specific performance with that of the top-performing team (TIAKong). As the results show, our method outperforms TIAKong for epidermis and blood vessel segmentation, while TIAKong delivers better results for the remaining tissue types on the final test set. 

In Table~\ref{tab:macro_f1_track1} and Table~\ref{tab:macro_f1_track2}, the nuclei-specific performance of our proposed approach, as well as that of the top-performing team (TIAKong) and the baseline model, is presented. As shown in Table~\ref{tab:macro_f1_track1} for Track 1, which includes three nuclei classes, our approach delivers superior performance for the Other class, while TIAKong achieves slightly better results for the Tumor and Lymphocyte classes. Compared to the baseline model, our approach outperforms it across all three nuclei classes. As depicted in Table~\ref{tab:macro_f1_track2} for Track 2, our method outperforms TIAKong for apoptosis, epithelium, histiocyte, neutrophil, and stroma, while TIAKong achieves better performance for the remaining nuclei types on the final test set. Once again, compared to the baseline model, our approach delivers superior performance across all ten classes by a large margin.


\begin{table*}
	\centering
	\caption{Tissue-based scores for Track 1 and Track 2 of the challenge on the final test set, based on the micro Dice score (\%).}
	\label{tab:tissue_deatils}
	\resizebox{\textwidth}{!}{%
	\begin{tabular}{lccccccr}
		\hline
		Track      & Team    & Tumor         & Stroma        & Necrosis       & Epidermis     & Blood vessel   & Mean \\
		\hline
		Track 1    & LSM      & 92.07         & 81.28         & 46.79          & \textbf{87.32} & \textbf{54.37} & 72.36 \\
		Track 2    & LSM      & 92.07         & 81.28         & 74.49          & \textbf{87.32} & \textbf{54.37} & 77.98 \\
		Track 1/2  & TIAKong  & \textbf{93.57} & \textbf{83.59} & \textbf{82.03} & 86.25          & 45.69          & \textbf{78.23} \\
		\hline
	\end{tabular}%
	}
\end{table*}

\begin{table}[ht]
\centering
\caption{Macro F1 scores of Track 1 for tumor, lymphocyte, and other nuclei classes (\%).}
\label{tab:macro_f1_track1}
\begin{tabular}{lcccc}
\hline
\textbf{Team} & \textbf{Tumor} & \textbf{Lymphocyte} & \textbf{Other} & \textbf{Macro F1} \\
\hline
LSM       & 82.87 & 79.24 & \textbf{61.17} & \textbf{74.43} \\
TIAKong   & \textbf{83.86} & \textbf{81.04} & 58.26 & 74.39 \\
Baseline  & 77.68 & 69.40 & 51.25 & 69.40 \\
\hline
\end{tabular}
\end{table}

\begin{table*}[ht]
\centering
\caption{Macro F1 scores of Track 2 for each nuclei class (\%).}
\label{tab:macro_f1_track2}
\begin{tabular}{lccc}
\hline
\textbf{Class} & \textbf{LSM} & \textbf{TIAKong} & \textbf{Baseline} \\
\hline
Apoptosis      & \textbf{38.00} & 19.46   & 5.86     \\
Endothelium    & 43.93 & \textbf{44.03} & 23.92     \\
Epithelium     & \textbf{75.35} & 60.83   & 3.07     \\
Histiocyte     & \textbf{45.66} & 43.22   & 36.11     \\
Lymphocyte     & 75.34   & \textbf{80.90} & 69.81     \\
Melaphonage    & 38.00   & \textbf{44.48} & 29.52     \\
Neutrophil     & \textbf{29.39} & 26.97   & 16.96     \\
Plasma         & 21.14   & \textbf{25.00} & 10.15     \\
Stroma         & \textbf{40.56} & 38.13   & 29.07     \\
Tumor          & 82.29   & \textbf{83.83} & 73.20     \\
\textbf{Macro F1}   & \textbf{48.96} & 46.69   & 29.77     \\
\hline
\end{tabular}
\end{table*}

\begin{figure}[h]
	\centering
	\includegraphics[width=1.0\textwidth]{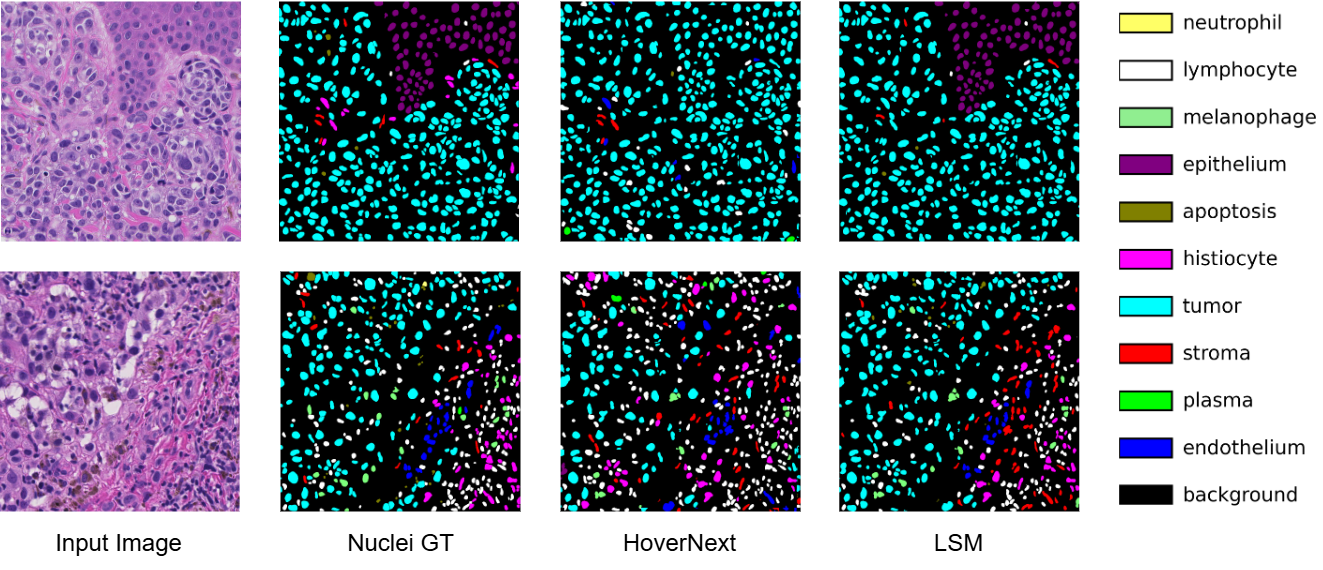}
	\caption{Example of nuclei segmentation and classification results from the PUMA dataset. GT: Ground Truth.}\label{fig:example}
\end{figure}

For visualization of nuclei segmentation, Figure~\ref{fig:example} presents a sample result from our method compared with the baseline HoVer-NeXt and the ground truth segmentation mask. As the qualitative results demonstrate, our predicted segmentation is notably closer to the ground truth than the baseline.

\begin{figure}[h]
	\centering
	\includegraphics[width=.60\textwidth]{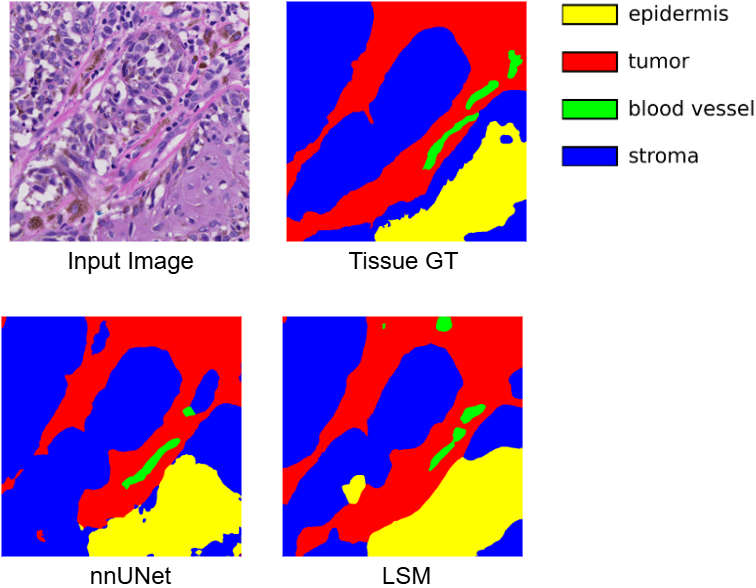}
	\caption{Example of tissue segmentation and classification results from the PUMA dataset. GT: Ground Truth.}\label{fig:exampleT}
\end{figure}

For tissue segmentation, Figure~\ref{fig:exampleT} presents an example result from our method alongside the baseline nnU-Net for comparison, where our model delivers qualitatively superior results.


\subsection{MoNuSAC Dataset Results}
As explained in Section~\ref{sec:Training Procedure for MoNuSAC}, we used the instance nuclei segmentation results from the MoNuSAC challenge winning team (TIA-Lab) and the post-challenge winning team (AmirrezaMahbod or AM) to investigate the impact of integrating the proposed multi-stage and auto-context approach on their performance. The results of these comparisons are presented in Table~\ref{tab:monusac}. As the nuclei-specific and average nuclei scores (mean Micro PQ and mean PQ) indicate, our proposed approach yields consistent improvements across all metrics for both baseline methods, with the exception of the macrophage class in the AM approach. These improvements suggest that leveraging global tissue-level information enhances the model’s ability to distinguish between different nuclei types, particularly in complex histological environments. It is worth mentioning that, for generating tissue segmentation masks in the MoNuSAC dataset, we used the PUMA tissue segmentation pre-trained model without any additional annotations or training from the MoNuSAC dataset.



\begin{table}[ht]
	\centering
	\caption{Comparison of class-specific and average micro Panoptic Quality (PQ) scores, as well as mean PQ, on the MoNuSAC test set. 
    }
	\label{tab:monusac}
	\resizebox{\textwidth}{!}{%
	\begin{tabular}{lcccccc}
		\hline
		Method & Epithelial & Lymphocyte & Neutrophil & Macrophage & Mean Micro PQ & Mean PQ \\
		\hline
	TIA-Lab               & 70.52          & 70.04          & 68.02          & 37.52           & 61.52            & 61.04   \\
	TIA-Lab with LSM      & \textbf{71.15} & \textbf{70.54} & \textbf{69.86} & \textbf{39.26}  & \textbf{62.71}   & \textbf{62.29} \\ \hline
        AM                    & 67.23          & 68.63          & 62.74          & \textbf{52.58}  & 62.82            & 59.57   \\
	AM with LSM           & \textbf{68.71} & \textbf{69.53} & \textbf{68.01} & 50.74           & \textbf{64.22}   & \textbf{62.18}   \\
		\hline
	\end{tabular}%
	}
\end{table}

\begin{figure}[h]
	\centering
	\includegraphics[width=1\textwidth]{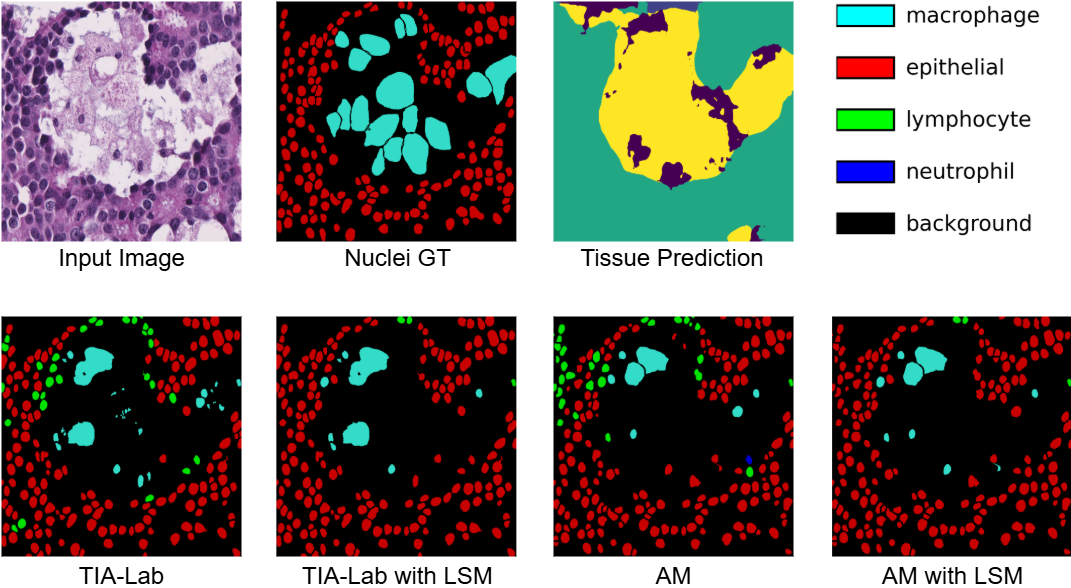}
	\caption{Example nuclei segmentation and classification results from the MoNuSAC dataset. The figure compares ground truth annotations with predictions from the MoNuSAC challenge winner (TIA-Lab), the post-challenge winner (AM), and the proposed method (LSM). The proposed method benefits from tissue context integration, resulting in improved nuclei type classification.    }\label{fig:exampleM}
\end{figure}

An example result from the MoNuSAC dataset is illustrated in Figure~\ref{fig:exampleM}. The inclusion of tissue-level information improves the performance of nuclei classification. As shown in the figure, falsely recognized lymphocytes in TIA-Lab and AM methods are correctly classified as epithelial cells when LSM method is additionally integrated.


\subsection{Ablation Study}

To better understand the contribution of individual stages to the overall performance of the proposed framework, we conducted an ablation study by systematically removing key components of the model. Specifically, we selected six components and designed seven experimental configurations by excluding these elements. All ablation experiments were performed using the internal validation set to assess the performance impact of each component.

\paragraph{Experiments 1 and 2: Effect of the Classifier}
To evaluate the impact of the classifier, we removed Stage 2 entirely and directly used the tissue segmentation output from the SegFormer model in Stage 1. As shown in the Exp1 and Exp2 rows of Table~\ref{tab:tissue_ablation}, this modification resulted in decreased segmentation performance for both tissue and nuclei. These results highlight the importance of incorporating a dedicated classifier, especially for tasks involving small datasets.
\paragraph{Experiment 3: Effect of the Classification Rules}
As discussed in Stage 1, two classification rules are used: (1) if Epidermis is present in the segmentation, the image is classified as primary, and (2) if the number of primary pixels exceeds metastatic pixels, the image is classified as primary. In Experiment 2, only the second rule was applied, whereas in Experiment 3, both rules were used. As shown in Table~\ref{tab:tissue_ablation}, incorporating the first rule led to a slight improvement in tissue segmentation performance.
\paragraph{Experiment 4: Effect of the U-Net Branch}
The U-Net model was integrated into both Stage 2 and Stage 4 to enhance blood vessel tissue segmentation. Our findings indicate that U-Net outperforms SegFormer for detecting blood vessels. Incorporating U-Net improved segmentation performance for both tissue and nuclei (Table~\ref{tab:tissue_ablation}). Given that blood vessels are small structures, these results suggest that SegFormer performs better on classes with larger spatial dependencies compared to U-Net.
\paragraph{Experiment 5: Effect of Stage 4 (Tissue Refinement Using Nuclei Information)}
In Stage 4, nuclei segmentation results are used to refine tissue predictions. In Experiment 5, we included this refinement step to assess its impact. The results demonstrate a clear improvement in overall tissue segmentation performance (Table~\ref{tab:tissue_ablation}). This supports the effectiveness of an auto-context strategy, where the model leverages its own outputs to enhance performance.
\paragraph{Experiments 6 and 7: Effect of Ensemble Rules and Post-processing}
Finally, we evaluated the impact of the ensemble rules and post-processing techniques. As shown in Table~\ref{tab:tissue_ablation}, these modifications led to a slight improvement in tissue segmentation accuracy, further refining the performance of the proposed method. It is worth noting that, compared to the results on the final test set of the PUMA challenge, the effect of post-processing on the internal validation dataset has a smaller impact.

\begin{table}[ht]
\centering
\caption{Ablation study results showing the effect of each module on segmentation performance. The last row (Mean) is the average of Macro F1, Dice, and Micro Dice. Best values in each metric are highlighted in bold.}
\label{tab:tissue_ablation}
\resizebox{\textwidth}{!}{%
\begin{tabular}{lccccccc}
\toprule
\textbf{Component} & \textbf{Exp1} & \textbf{Exp2} & \textbf{Exp3} & \textbf{Exp4} & \textbf{Exp5} & \textbf{Exp6} & \textbf{Exp7} \\
\midrule
Classifier                & \xmark & \cmark & \cmark & \cmark & \cmark & \cmark & \cmark \\
Classification Rules      & \xmark & \xmark & \cmark & \cmark & \cmark & \cmark & \cmark \\
Unet Branch               & \xmark & \xmark & \xmark & \cmark & \cmark & \cmark & \cmark \\
Stage 4                   & \xmark & \xmark & \xmark & \xmark & \cmark & \cmark & \cmark \\
Tissue Ensemble Rules     & \xmark & \xmark & \xmark & \xmark & \xmark & \cmark & \cmark \\
Post Processing           & \xmark & \xmark & \xmark & \xmark & \xmark & \xmark & \cmark \\
\midrule
Macro F1                  & 31.47  & 32.26  & 32.26  & 33.88  & \textbf{33.88} & \textbf{33.88} & \textbf{33.88} \\
Dice                      & 74.06  & 81.08  & 81.09  & 80.04  & 82.05 & 81.93 & \textbf{82.11} \\
Micro Dice                & 63.44  & 72.99  & 73.00  & 74.91  & 77.65 & \textbf{77.76} & 77.66 \\
Mean                      & 56.32  & 62.11  & 62.11  & 62.94  & 64.52 & 64.52 & \textbf{64.55} \\
\bottomrule
\end{tabular}%
}
\end{table}

\section{Conclusion}

In this work, we introduced a novel multi-stage auto-context deep learning framework for tissue and nuclei segmentation and classification in H\&E-stained histological images of advanced melanoma. By leveraging contextual information across stages and integrating multiple state-of-the-art segmentation models, our method achieved excellent performance in the PUMA Challenge, ranking second in Track 1 and first in Track 2. Furthermore, the proposed approach outperformed the winners of the MoNuSAC challenge in terms of nuclei classification. These results highlight the effectiveness of fusing tissue and nuclei information within a unified framework.

\section*{Acknowledgment}
This work was supported by the Vienna Science and Technology Fund (WWTF) and by the State of Lower Austria [Grant ID: 10.47379/LS23006].


\bibliographystyle{elsarticle-num} 
\bibliography{refs.bib}

\end{document}